\documentclass[letterpaper, 10 pt, conference]{ieeeconf}  
\IEEEoverridecommandlockouts                              

\usepackage{microtype}
\usepackage{comment}

\usepackage{graphicx} 
\graphicspath{{./figures/}}
\usepackage{subcaption}
\usepackage{float}

\usepackage{amsmath} 
\usepackage{amssymb}  
\usepackage{units}
\usepackage{bm} 
\usepackage{multirow, array}

\usepackage{algorithm}
\usepackage{algpseudocode}

\makeatletter
\let\NAT@parse\undefined
\makeatother

\usepackage[square, numbers, sort&compress]{natbib}
\usepackage[hidelinks]{hyperref}
\usepackage[capitalise]{cleveref}

\crefname{equation}{}{}
\crefname{figure}{Fig.}{Figs.}

\usepackage{graphicx} 
\graphicspath{{./figures/}}
\usepackage{subcaption}
\usepackage{float}
\usepackage[font=small]{caption}
\usepackage{xparse}
\usepackage{xcolor}
\usepackage{stfloats}
\usepackage{diagbox}
\usepackage{makecell}
\usepackage{tabularx}
\usepackage{tabulary}

\newcommand\Tstrut{\rule{0pt}{2.6ex}}         
\newcommand\Bstrut{\rule[-0.9ex]{0pt}{0pt}}   
\newcommand\Bstrutbig{\rule[-1.5ex]{0pt}{0pt}}   

\title{\LARGE \bf Sim-to-Real Learning for Bipedal Locomotion Under\\Unsensed Dynamic Loads}
\author{\authorblockN{Jeremy Dao, Kevin Green, Helei Duan, Alan Fern, Jonathan Hurst}
\thanks{*This work is supported by the NSF Grant No. IIS-1849343, DGE-1314109, and DARPA Contract W911NF-16-1-0002.}
\thanks{All authors are with Collaborative Robotics and Intelligent Systems Institute, Oregon State University, Corvallis, Oregon, 97331, USA. }
\thanks{Email: \{\footnotesize daoje, greenkev, duanh, afern, jhurst\}@oregonstate.edu.} 
}

\begin{document}

\maketitle
\thispagestyle{empty}
\pagestyle{empty}

\begin{abstract}

Recent work on sim-to-real learning for bipedal locomotion has demonstrated new levels of robustness and agility over a variety of terrains. However, that work, and most prior bipedal locomotion work, have not considered locomotion under a variety of external loads that can significantly influence the overall system dynamics. In many applications, robots will need to maintain robust locomotion under a wide range of potential dynamic loads, such as pulling a cart or carrying a large container of sloshing liquid, ideally without requiring additional load-sensing capabilities. In this work, we explore the capabilities of reinforcement learning (RL) and sim-to-real transfer for bipedal locomotion under dynamic loads using only proprioceptive feedback. 
We show that prior RL policies trained for unloaded locomotion fail for some loads and that simply training in the context of loads is enough to result in successful and improved policies. 
We also compare training specialized policies for each load versus a single policy for all considered loads and analyze how the resulting gaits change to accommodate different loads. Finally, we demonstrate sim-to-real transfer, which is successful but shows a wider sim-to-real gap than prior unloaded work, which points to interesting future research. 


\end{abstract}
\section{Introduction}

The primary goal of robotics research is to develop practical robots that can act usefully in the real world.
To that end, there has been recent progress in bipedal locomotion research, which aims to allow robots to navigate environments built for humans.
There are few practical applications, however, that can be solved with just locomotion capabilities. In particular, it is important for robots to be able to handle and carry a variety of loads while maintaining robust locomotion.
Ideally, such load-carrying capabilities would not require additional sensing, allowing for easily swapping loads without needing additional hardware or wiring. 
Surprisingly, the problem of locomotion control while carrying such loads is under-researched, especially for bipedal robots.

In this work, we are particularly interested in \textit{dynamic loads}, such as an attached cart or container of liquid, rather than simple static loads like rigidly attached fixed masses. Such loads can have their own internal dynamics separate from the robot that can have effects even if the robot is stationary. Thus, locomotion with dynamic loads requires not only being robust to disturbances in the robot's base dynamics, but also controlling the external object via the locomotion gait.
These kinds of scenarios have additional underconstrained degrees of freedom that make the problem of locomotion much more complicated.
This raises the question of the effectiveness of recent learning-based approaches to bipedal control when extended to handle dynamic loads, especially without additional load-sensing capabilities.

\begin{figure}[t]
    \centering
    \includegraphics[width=0.95\columnwidth ]{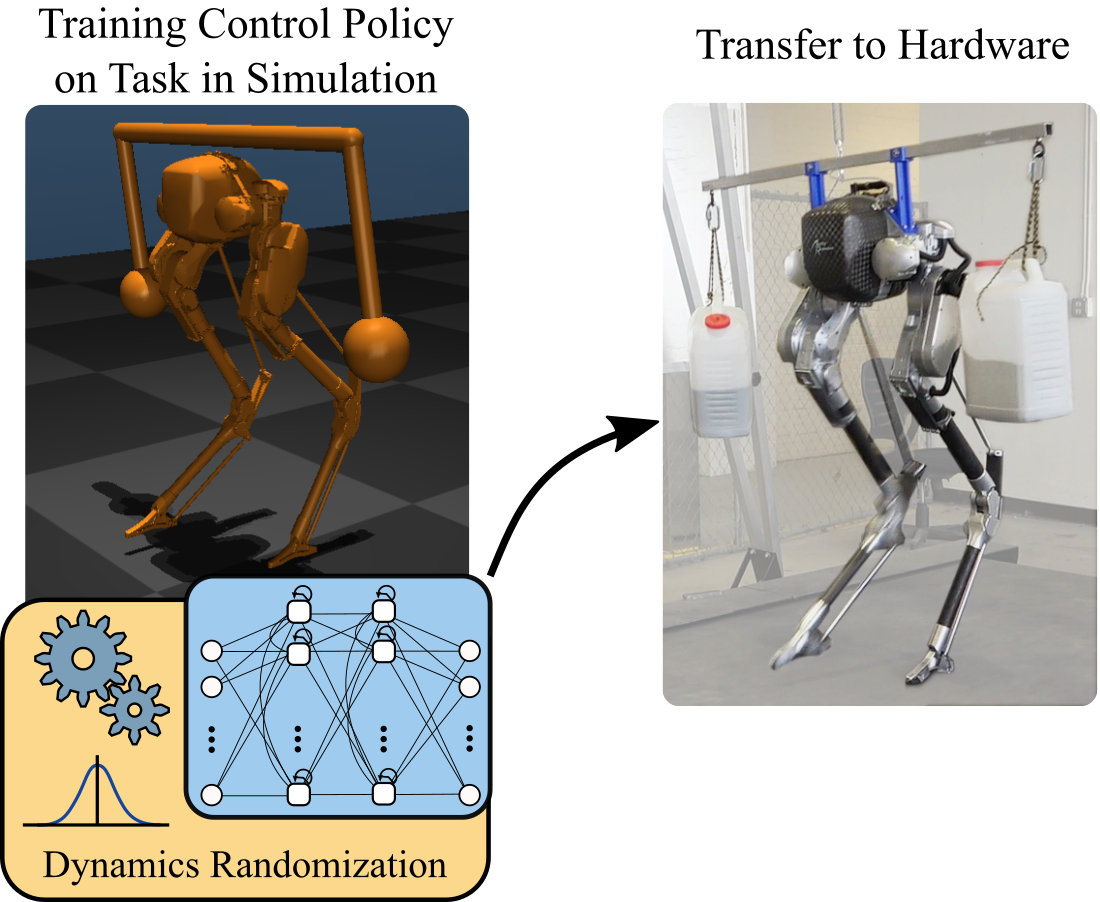}
    \caption{Cassie with two \unit[5]{kg} swinging masses attached. We are able to train policies capable of walking with such a dynamic load and transfer simulation trained policies to hardware.}
    \label{fig:cassiepole_hard}
\end{figure} 

One strategy for incorporating dynamic loads could be to use prior RL-trained controllers (for unloaded settings) to walk very slowly with simple gaits in order to avoid disturbing the load and having it impact the overall dynamics. Rather, our goal is to achieve similar levels of dynamic walking performance as these prior controllers but now with loads. To this end, we investigate the RL approach to learning locomotion policies for a number of qualitatively different dynamic loads with no additional load sensing beyond proprioception.

Our primary contribution is successful sim-to-real experiments of RL policies capable of handling varying dynamic loads on the bipedal robot Cassie. 
We show that simple straightforward training with specific loads is enough to result in effective locomotion that generally improves over prior base controllers trained without loads; no additional sensing capabilities, reward shaping, or modification to the nominal learning setup is required. 
We provide an analysis of the load-specific gaits compared to the base gaits to understand the key strategic differences. In addition, we find that a general load policy trained on all of the loads is able to reach nearly the same walking performance as the load-specific policies. We also show that training load policies can be dramatically accelerated by starting with a pre-trained unloaded policy. Finally, we evaluate sim-to-real performance of the learned load policies. We show that the policies do successfully transfer to the real-world, however, the sim-to-real gap is larger than in previous unloaded work, which suggests interesting future research opportunities.

\section{Related Work}
\subsection{RL for Bipedal Locomotion}
There have been many recent successes of RL applied to bipedal locomotion. 
Many of the methods have relied on some kind of expert data in the reward function in order to properly inform the policy how to walk. 
This is usually in the form of trajectories \cite{Peng2017, Yu2019} or motion capture data \cite{Peng2018a,Yang2020a, Taylor2021}.
Other works have moved away from the use of such data to instead use a principled reward function that describes walking by alternating between rewarding swing and stance of each foot \cite{Siekmann2020}. 
This allows the reward for a walking gait to be described through simple gait parameters like swing ratio and cycle time. 
The most common method to transfer simulator-trained RL policies to hardware is via dynamics randomization \cite{Peng2018, deepmind_minitaur, Mozifian2019}, where the policy is trained over a distribution of simulation model parameters to improve robustness to inaccuracies in the simulator. In this work, we use a reward structure and dynamics randomization similar to prior sim-to-real demonstrations on the Cassie robot \cite{Siekmann2020}.

\subsection{Locomotion with Loads}

Though there are many studies in the field of loco-manipulation \cite{Wu2019, Saputra2018, Wu2021}, \textit{legged} loco-manipulation has received limited attention. 
Previous research on quadrupeds with an attached manipulation arm \cite{Bellicoso2019, Sleiman2021, Tan2020, Wolfslag2020} rely mainly on the statically stable nature of having four legs and mainly focus on holding static weights. 
For example, recent works \cite{Bellicoso2019, Sleiman2021} optimize center of mass and limb motions which are tracked with a whole body controller. 
However, this requires the load to be fixed to the robot since the whole body controller assumes that the load is just another rigid part of the robot. 
    
Study of loco-manipulation in the context of bipedal robots has largely separated out the locomotion and manipulation problem. 
For example, \cite{Settimi2016} and \cite{Ferrari2017} investigate the problem of walking to an object, then picking it up or moving it while stationary, not carrying a load \textit{while} locomoting as we are concerned with here. 
\cite{Vaz2020, Vaz2017} do target walking with a load, but use statically stable walking gaits which limits the dynamic effects of the load and results in slow walking speeds.
\cite{Vaz2020a} had a human scale DRC-Hubo robot climb stairs while holding half-full 2 gallon buckets of water in each hand. 
This was achieved by using an approximate model of the sloshing dynamics and a PID controller feeding back on the joint angles, wave height of the container liquid, and force data of the wrist, to adjust the arm joint angle. 
It is noted that this method does use sensing not required for walking (wrist force data and liquid wave height) and that the load's force was ``not enough to change the center of gravity of the robot in an unstable manner," so only adjustment of the arm motion was required to be successful. In contrast, we focus on the feasibility of learning to carry such dynamic loads without specialized sensing and control architectures.
\section{Learning Load Policies}

Our learning approach is based on a prior RL framework for training bipedal locomotion policies \cite{Siekmann2020} in an unloaded setting. Here we first summarize the key elements of the framework, referring the reader to the original paper for further details. Next, we describe our adaptations needed for extending to locomotion under dynamic loads.

\subsection{Locomotion Learning Framework}

Following prior work \cite{Siekmann2020} we use an LSTM neural network as our primary locomotion controller with two 128-dimensional recurrent hidden layers. The network input contains: 1) the robot state information, including pelvis orientation, rotational velocity, and joint positions and velocities, 2) a clock signal consisting of a sine and cosine function with period length equal to the desired cycle time, and 3) the desired forward velocity. The network outputs position set points for each joint at 40Hz, which are passed to a PD controller with fixed gains running at 2kHz. The network is trained in simulation via the actor-critic PPO algorithm \cite{Schulman2017} using the clipped objective along with gradient clipping. The simulation environment is based on the \textit{cassie-mujoco-sim} \cite{AgilityRobotics2018} library and the MuJoCo \cite{todorov2012mujoco} physics engine. To facilitate sim-to-real transfer, we use dynamics randomization in our training where we randomize joint damping, joint mass, and ground friction following prior work \cite{Siekmann2020}.

\begin{figure}
    \centering
    \vspace{0.1cm}
    \includegraphics[width=0.48\textwidth]{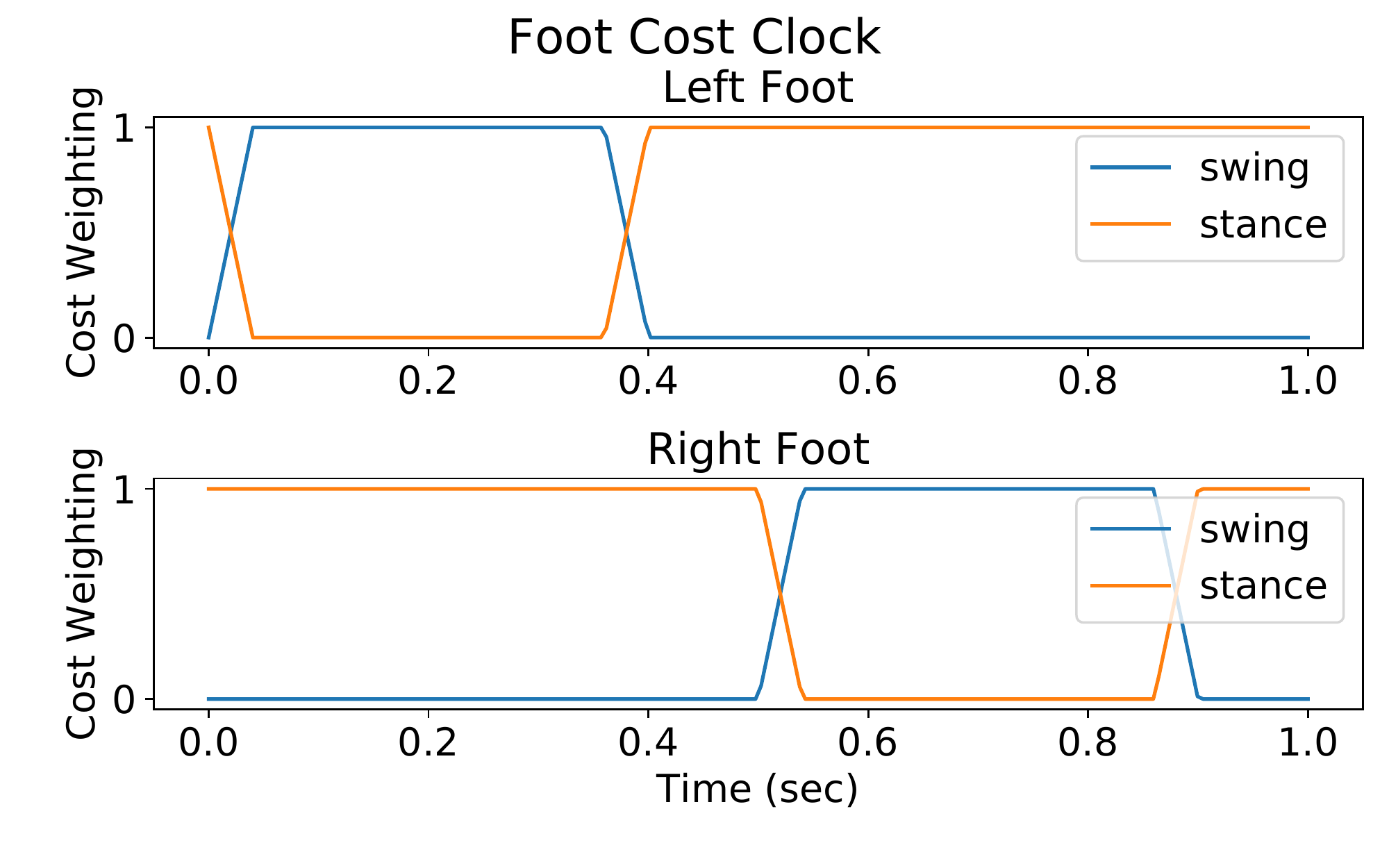}
    \vspace{-.85cm}
    \caption{An example of the piecewise-linear foot weighting clock with 40\% swing ratio and cycle time of 1 sec. The foot will be in swing for 0.4 sec, 20\% of which is used to transition linearly between 0 and 1 and back again. Weighting for stance is just 1 - swing weight. To get the weightings for the right foot we just shift the left foot function forward half a cycle or 0.5 seconds.}
    \vspace{.25cm}
    \label{fig:linclock}
\end{figure}

The reward signal for training is constructed the same as in \cite{Siekmann2020} consisting of the error between commanded velocity/orientation and actual velocity/orientation, smoothing terms like pelvis acceleration penalty, torque penalty, and an action difference cost, and foot costs to encourage alternating swing and stance of each foot. However, to define the foot stance and swing cost weightings, we replace the probabilistic function used in \cite{Siekmann2020} with a simple piecewise linear periodic function. In particular, our foot cost weighting clock function is a set of anchor points defined by the swing ratio and total cycle time, between which we linearly interpolate. \cref{fig:linclock} shows an example linear clock with a swing ratio of 40\% and a cycle time of 1 second.



Since, in this work, we focus on walking at different speeds, rather than covering the entire gait spectrum as in prior work \cite{Siekmann2020}, we further simplify the learning by only training on a fixed swing ratio and stepping frequency for each speed. The phase offset gait parameter is always set to 0.5 for symmetrical walking. We train on speeds from \unit[0]{m/s} to \unit[4]{m/s} and \cref{table:swing_step} shows the heuristically defined swing ratio and stepping frequency for each speed. For speeds \unit[1]{m/s} to \unit[3]{m/s} we linearly increase the swing ratio from 0.4 to 0.8 and stepping frequency from \unit[1]{Hz} to \unit[1.5]{Hz} in relation to the commanded speed. 

\begin{table}
\centering
 \begin{tabular}{|c | c| c|} 
 \hline
 Speed Range [\unit{m/s}] & Swing Ratio [\%] & Step Frequency [\unit{Hz}] \\ 
 \hline
 0 - 1 & 40 & 1 \\ 
 \hline
 1 - 3 & 40 - 80 & 1 - 1.5 \\
 \hline
 3 - 4 & 80 & 1.5 \\
 \hline
\end{tabular}
\caption{Heuristics of gait parameters used in our running policy. }
\label{table:swing_step}
\end{table}

Using the above method for training in simulation without loads results in a robust and stable walking policy that can accept velocity and orientation commands. We will refer to this policy as the \textit{base policy}, which will be used as a point of comparison in our experiments and for bootstrapping learning of load policies. 

\subsection{Learning Locomotion for Dynamic Loads}

\begin{figure*}[t]
    \centering
    \vspace{0.3cm}
    \includegraphics[width=0.85\textwidth]{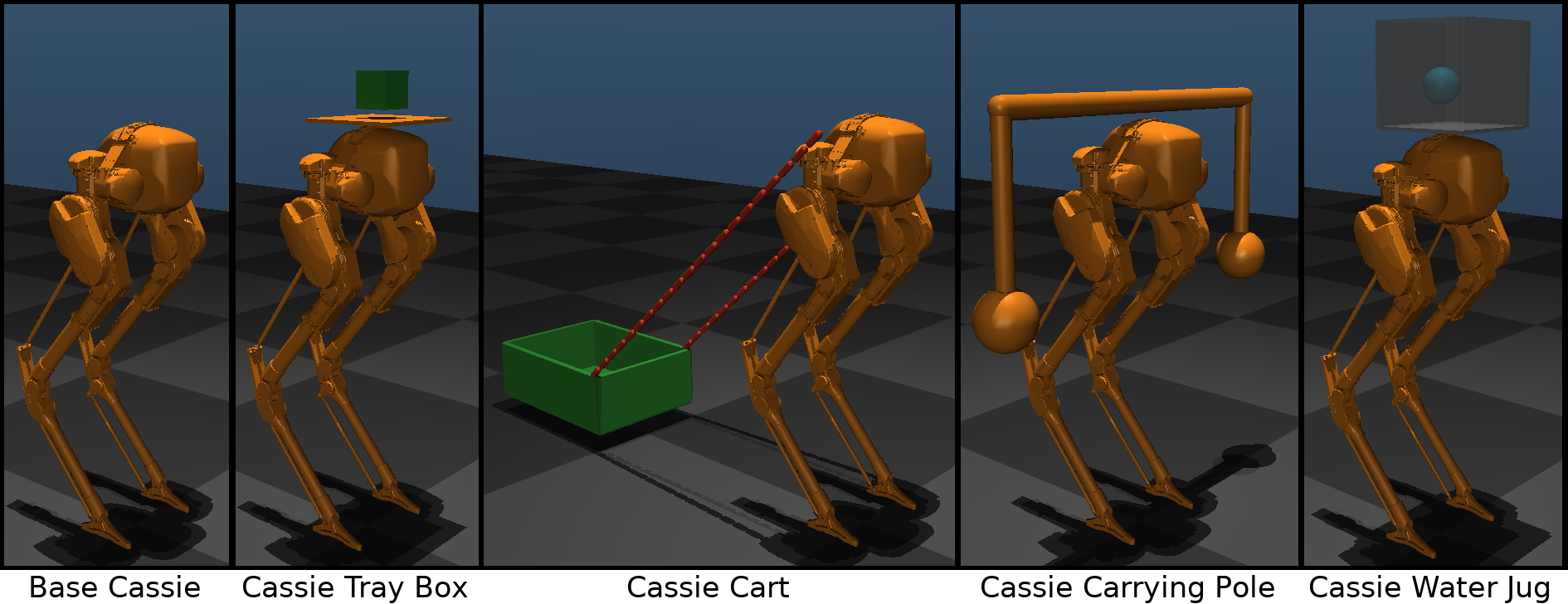}
    \caption{All the different Cassie models that we experiment on, showing the 4 different dynamic loads as well as the base Cassie model.}
    \label{fig:all_loads}
    \vspace{-.6cm}
\end{figure*} 

Training policies to handle dynamic loads requires minimal modification to our base training setup. The primary modification is contained to the simulator where we define and use loaded robot models instead of the regular unloaded model during simulations. The only exception is in cases where the reward function might be modified to have a component related to a desirable load property (see below), e.g. to punish spilling a liquid. 
%

In this work, we consider 5 qualitatively different loaded Cassie robot models, chosen to test different kinds of dynamic effects and possible use cases.
\begin{itemize}
    \item \textbf{Unloaded Cassie} model, the regular Cassie model that has no load attached and that the base walking policy is trained from.
    \item \textbf{Cassie Tray Box}, which consists of a freely moving \unit[5]{kg} box on top of a tray attached to the top of the pelvis. The policy must learn to handle a free external object such that it does not drop, similar to a waiter carrying a cup.
    \item \textbf{Cassie Cart}, which consists of a cart attached with ropes to the pelvis of Cassie dragged along the ground. This load is used to test backwards pulling forces.
    \item \textbf{Cassie Carrying Pole}, which consists of a carrying pole attached to the pelvis with hanging \unit[5]{kg} weights on each end. Since the weights are free to swing in every direction, this difficult load can easily cause both lateral and torsional forces that induce unstable cycles and destroy locomotion. 
    \item \textbf{Cassie Water Jug}, which consists of a box rigidly attached to the pelvis with water inside. To mimic the liquid dynamics without having to do fluid simulations, we approximate the sloshing dynamics with a weight on springs in every axis. 
\end{itemize}

For each of these loads, expect for Tray Box, we used the same reward function as described in Section II A that simply encourages smooth locomotion at a desired velocity. 
For the Tray Box load, since there is the additional goal of keeping the box on top of the tray, we add an extra reward component that penalizes the distance between the position of the box and middle of the tray.
We additionally encourage the policy to keep the box on the tray by using an extra termination condition. 
In addition to the reward and height termination conditions described in \cite{Siekmann2020} we terminate a trajectory if the box's z height falls below \unit[50]{cm}, effectively meaning the box has fallen off the tray.




Finally, we experiment with three ways of training load policies. 
First, we consider training load-specific policies from \textit{scratch} starting from a random neural network.
Second, we consider training load-specific policies by \textit{bootstrapping} from the base policy, which simply initializes learning with the high-performing policy trained without loads.
This is similar to the type of pre-training of general models in computer vision and natural language processing, followed by fine-tuning for specific applications.
Finally we consider, training a general load policy that is trained on all load models equally.
This means that for each training iteration, the 50000 timesteps sampled are split evenly between each of the 5 models, ensuring that the policy sees equal amounts of experience from each load.
Note that in this case, since we cannot train the policy on multiple different reward functions, we use the base walking reward function even for the Tray Box environment unlike in the load specific policies.
The ``box falling off" termination condition is kept for trajectories gathered in the tray box model. We train both from scratch and bootstrapped versions of this general load policy.

\section{Simulation Results}
We consider the following metrics to evaluate walking performance under dynamic loads in simulation. 

\textbf{Pass Rate}. We randomly sample 5000 command sets, which consist of 2 random speed commands and two random orientation commands, and measure how many sets it can successfully follow. New commands are applied every 2.5 seconds. To ensure some minimum change, the speed commands are randomly chosen between 0.5 and \unit[2.0]{m/s} higher or lower than the previous command (within the command limits of 0 to \unit[4]{m/s}) and the orientation change command is similarly between $30^{\circ}$ to $60^{\circ}$ left or right. 
A trial is considered a failure if the robot falls down at any point.

\textbf{Average Speed Error}. For command set cases where the robot does not fall down, we additionally track the average difference in desired and actual robot velocity over the trajectory. We do not report this measure when the pass rate is less than 50\%.

\textbf{Average Force}. To evaluate push perturbation robustness, we perform a line search in each direction to find the maximum force that the policy is able to resist. The policy gets 3.0 seconds to recover from the 0.2 second impulse that is applied to the robot pelvis. If the robot has not fallen down after the 3 seconds, it is considered successful. We test every $10^{\circ}$ and start from \unit[50]{N}, increasing by \unit[10]{N} until the force is too much and the falls over. We then average over each direction to get a single number that gives us an idea of how much force the policy can resist.

\textbf{Max Speed}. We test what is the maximum speed the policy is able to walk at without falling over.


\subsection{Base vs. Specialized Policies}

In all load cases, we are able to successfully train specialized policies that are able to walk with the load. 
\cref{table:load_walk_comp} shows the walking performance for the base walking policy evaluated on each different model, as well as the performance of each load specific policy. 
Each policy was trained from scratch.
We can see how the performance of the base policy degrades when evaluated on the dynamic loads it has never seen before, and how much performance the load specific policies are typically able to gain back.

\textbf{Pass Rate.} In almost all cases the base policy fails on more command sets than the load specific policies. 
This is especially true of the challenging Carry Pole load, on which the base fails nearly immediately. 
Just by training on the Carry Pole model, we are able to get an over 95\% pass rate improvement, going from falling down to stable and responsive walking. 


\textbf{Average Force Resistance.} In all cases the load specific policies outperform the base policy, being able to resist 10-20N more force on average. 
Again, we see the largest improvement in the Carry Pole case, where force resistance more than doubles. 
It is also interesting to note that some of the easier loads actually have a stabilizing effect on the gait. 
For example, when evaluated on the Cart load which is has never seen the before, the base walking policy can actually resist \unit[60]{N} more force on average than when evaluated on the unloaded model. 
This is likely due to Cassie's morphology, particularly its lack of any arms or torso. 
This lack of inertia can make it hard to resist twisting forces. 
Adding loads increases the overall inertia of the robot and depending on the load can actually do so in a beneficial way like with the cart load.

\textbf{Average Speed Error.} The load specific policies achieve lower average error compared to the base policy on the Cart and Water Jug loads, which are the only loads where the base policy has high pass rates and hence meaningful speed errors. Interestingly, we see that the load specific policy for the Water Jug achieves an error that is less than the base policy in the unloaded condition. This is another example of unintended positive impacts of loaded conditions.


\begin{table}[h]
\centering
\begin{tabularx}{\columnwidth}{|c|c|c|X|c|X|c|} 
\hline
 & &\thead{Unloaded} & \thead{Tray\\Box} & Cart & \thead{Carry\\Pole} & \thead{Water\\Jug}\\
\hline
\multirow{3}*{\thead{Pass\\Rate}}\Tstrut\Bstrut & Base\Tstrut\Bstrut & .9784 & .5364 & .9998 & .0014  & .9746\\ \cline{2-7}
 & Specific\Tstrut\Bstrut & .9784 & .7112 & .996 & .9578 & .976\\[1pt] \cline{2-7}
 & General\Tstrut\Bstrut & .9992 & .91 & 1.0 & .8342 & .9998\\[1pt]
\hline
\multirow{3}*{\thead{Avg\\Force\\(\unit{N})}}\Tstrut\Bstrut & Base\Tstrut\Bstrut & 168.89 & 213.89 & 208.89 & 95 & 184.17\\ \cline{2-7}
 & Specific\Tstrut\Bstrut & 168.89 & 236.39 & 218.89 & 221.11 & 202.22\\[5pt] \cline{2-7}
 & General\Tstrut & 180.83 & 205.56 & 233.61 & 189.72 & 190.28\\[5pt]
\hline
\multirow{3}*{\thead{Avg\\Speed\\Error\\(\unit{m/s})}} & Base\Tstrut\Bstrutbig & .242 & .152 & .3396 & N/A & .257\\[2pt] \cline{2-7}
 & Specific\Tstrut\Bstrut & .242 & .175 & .3011 & .3887 & .2353\\[2pt] \cline{2-7}
 & General\Tstrut\Bstrut & .3182 & .2915 & .4463 & .437 & .3261\\[2pt]
\hline
\multirow{3}*{\thead{Max\\Speed\\(\unit{m/s})}} & Base\Tstrut\Bstrut & 3.9 & 1.8 & {3.5} & N/A & {3.6}\\ \cline{2-7}
 & Specific\Tstrut\Bstrut & {3.9} & {2.1} & 2.6 & {2.5} & 3.0\\ \cline{2-7}
 & General\Tstrut & 2.6 & 1.9 & 2.4 & 2.5 & 2.6\\
\hline
\end{tabularx}
\caption{Walking performance comparison between the base policy, the specialized policies, and the general load policy. The specialized policies are only evaluated on the model they were trained on while the base and general load policy are evaluated on every model. All policies are trained from scratch.}
\label{table:load_walk_comp}
\vspace{-.2cm}
\end{table}

\textbf{Max Speed} Something that was unexpected is that for the Cart and Water Jug load the base policy has a higher max speed than the policy specifically trained on the loads. 
This means that a higher walking speed is possible, but for some reason a policy trained on the load won't learn such a gait. 
We hypothesize that this is because of the termination conditions which greatly emphasizes staying alive over strictly following the speed command. 
This may cause the load specific policies to be more conservative than the base policy, which doesn't see any load during training and likely triggers the termination condition much less.

Though we are unable to gain back full performance of the base policy in some cases, this is expected as adding in a load fundamentally changes the problem and possibly changes the walking performance limit. It is unreasonable to assume that a policy could walk as fast as before attaching swinging weights to the robot.

\subsection{Specialized vs. General Load Policies}
We are also able to learn a general load policy that can handle a number of different loads. 
It is reasonable to assume that the skills needed to handle one load can be applied to a different load, and that similar sorts of stabilizing, minimal movement gaits are needed no matter the load morphology. 

\cref{table:load_walk_comp} shows the walking performance of the general load policy versus each model's specifically trained policy. 
In most cases performance between the two is comparable, though the general load policy does have higher speed error and on the harder Tray Box and Carry Pole loads handles around \unit[30]{N} less push force. 
This indicates that a general load policy is viable, and that there is perhaps some general load robustness skill that can be learned. 

One case where the general load policy makes a significant compromise is max speed for the unloaded condition. 
Since the policy has to handle walking for all loads, the policy never gets the chance to explore higher speed walking as it would cause failures on the load models. Thus even when evaluated in the unloaded condition the policy can only reach 2.6 m/s. We also see a significant decrease in pass rate for the Carry Pole load compared to the specialized policy, which is arguable the most difficult dynamic load in our set. While the general purpose load policy is much better than the base policy, which has never experienced loads, this is a good example of when specialized behavior is still beneficial. Finally, it is interesting that the general policy typically improves the pass rate on other loads compared to the base and specialized policies. This suggests that it has gained robustness due to needing to handle a variety of dynamic conditions.


\subsection{Learning from Scratch vs. Bootstrapping}
We also investigate bootstrapping the learning from a base unloaded walking policy compared to learning each load specific policy from scratch.
Since in every case the goal is still to walk forward, it makes sense that an expert walking policy would have useful information and skills for the load policies, and that we can quickly generalize from such an expert policy to learn such a similar walking task. 

\begin{figure*}[t]
    \centering
    \vspace{.1cm}
    \includegraphics[width=\textwidth]{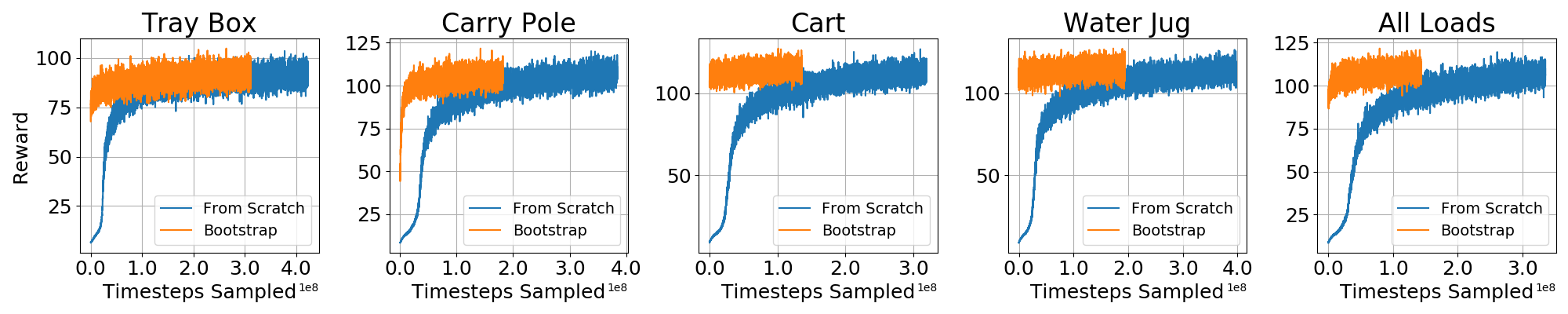}
    \caption{Learning curve comparison between learning from scratch and bootstrapping from a successful walking policy for each load type. Note how in all cases bootstrapping is significantly more sample efficient.}
    \label{fig:rew_compare}
    \vspace{-.5cm}
\end{figure*}

We are able to learn bootstrapped policies for all loads without catastrophic unlearning, even in the case of the carry pole load on which the base walking policy completely fails.
Even though the reward starts out drastically low as the robot continually falls down, the base walking information embedded in the policy network is still useful and the policy is quickly able to learn to walk again.
\cref{fig:rew_compare} shows the learning curves between the bootstrapped policies and the ones learned from scratch.
As we expected, the bootstrapped policies need on the order of 100 million less samples to reach the same reward.
 
\cref{table:load_walk_comp_bootstrap} shows that both types of policies have similar walking performance and in some cases, even higher maximum speed and lower average speed error. This means that we can bootstrap without worry of a loss in performance.

\begin{table}
\centering
\vspace{.1cm}
\begin{tabularx}{\columnwidth}{|c|c|X|X|X|X|} 
\hline
 & & \thead{Tray\\Box} & Cart & \thead{Carry\\Pole} & \thead{Water\\Jug}\\
\hline

\multirow{2}*{Pass Rate}&Scratch\Tstrut\Bstrut & .7112 & .996 & .9578 & .976\\ \cline{2-6}
 & Bootstrap\Tstrut\Bstrut & .6944 & .9986 & .9834 & .998\\
\hline

\multirow{2}*{Avg Force (\unit{N})} & Scratch\Tstrut\Bstrut & 236.39 & 218.89 & 221.11 & 202.22\\ \cline{2-6}
 & Bootstrap\Tstrut\Bstrut & 239.44 & 189.44 & 226.67 & 192.78\\
\hline
\multirow{2}*{\thead{Avg Speed\\Error (\unit{m/s})}} & Scratch\Tstrut\Bstrut & .175 & .3011 & .3887 &  .2353\\[1pt] \cline{2-6}
 & Bootstrap\Tstrut\Bstrut & .1434 & .2111 & .3533 & .2156\\[1pt]
\hline
\multirow{2}*{Max Speed (\unit{m/s})} & Scratch\Tstrut\Bstrut & 2.1 & 2.6 & 2.5 & 3.0\\ \cline{2-6}
 & Bootstrap\Tstrut\Bstrut & 2.0 & 3.2 & 2.4 & 3.1\\
\hline
\end{tabularx}
\caption{Walking performance comparison between learning a policy from scratch versus bootstrapping from an expert walking policy.}
\label{table:load_walk_comp_bootstrap}
\end{table}

\subsection{Gait Analysis}


By comparing phase portraits between the base policy and each load specific policy, we can examine what strategies the policy learns to deal with each load.

\begin{figure}
    \centering
    \includegraphics[width=0.98\columnwidth]{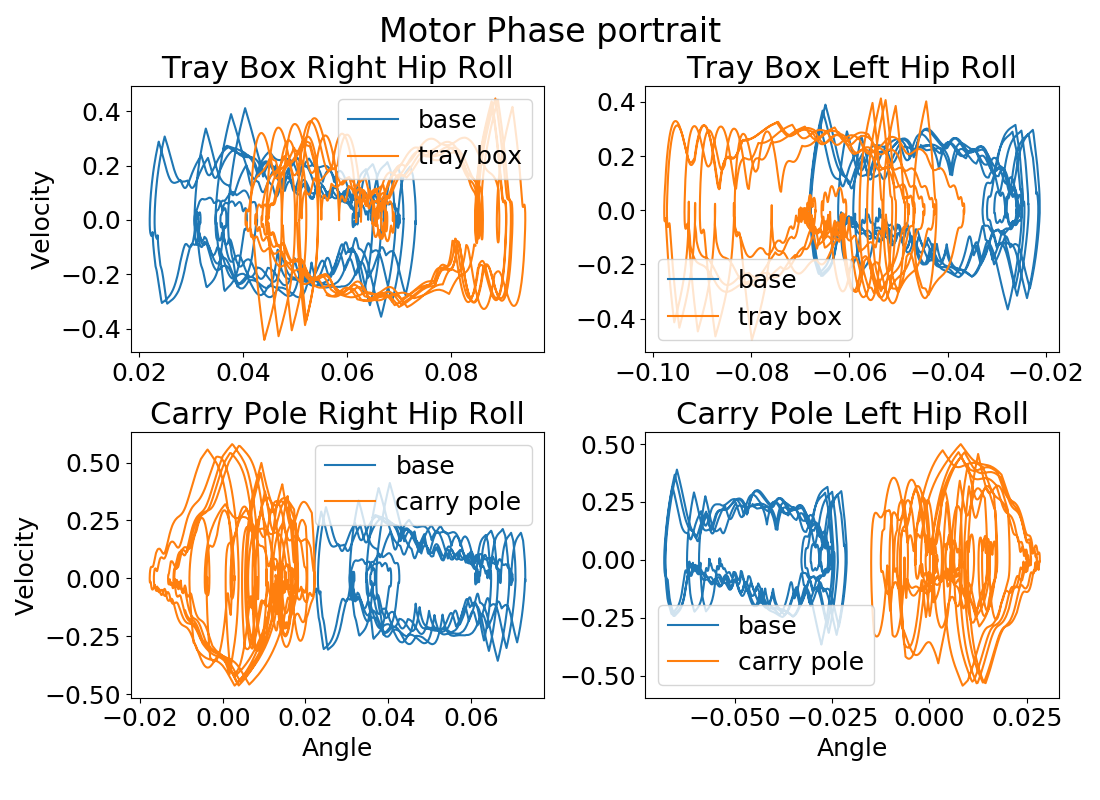}
    \vspace{-.3cm}
    \caption{Phase portrait comparison of the hip roll joints for the base policy vs. the Tray Box policy and base policy vs. the Carry Pole policy. Note the difference in walking strategy, as the tray box policy takes narrower steps and the carry pole policy takes wider steps.}
    \label{fig:portrait_comp}
\end{figure}

For example, \cref{fig:portrait_comp} compares the phase portraits of the tray box and base policy in the top half. 
We show only the hip roll joints since it is the area of main difference and the other actuated joints fall into similar cycles. 
Note that the signs for the left and right hip are opposite because they are on opposite sides of the pelvis; for the left roll negative angle means towards the pelvis and for the right roll positive angle means towards the pelvis. 
Here, the Tray Box policy in orange has noticeably more positive right roll values and more negative left roll values. 
This indicates that the Tray Box policy's legs are kept closer towards the center of the pelvis. 
In other words it's taking narrower steps to help keep the pelvis more stable since a narrower step path means the pelvis has to move less as the weight of the robot shifts to be more under the foot currently in stance.
A more stable pelvis results in less movement in the box and the box not falling off the tray. 


We can see that this is a different strategy from the Carry Pole policy.
The bottom half of \cref{fig:portrait_comp} shows the phase portrait for the Carry Pole policy. 
Looking again at the hip roll values, we see that the left values are more positive and the right values are more negative.
This indicates the exact opposite strategy of the tray box policy: the Carry Pole policy takes wider steps than normal instead of narrower. 
This is likely because the swinging masses exert much larger forces on the robot than the loose box, and thus the policy needs to walk with a wider stance in order to resist the extra lateral and torsional forces applied by the swinging weights.

\section{Hardware Results}

We test the tray box and carry pole policies on hardware in addition to the base and general load policy. 
All policies are able to successfully transfer to hardware. 
We refer readers to following link to view the hardware results: \url{https://youtu.be/IeSUM_ej8wE.} We conduct our hardware tests by starting the policy walking in place, then slowly increasing the speed until the robot falls down or the policy can not make the robot walk any faster. 
The successful policies are able to walk without falling down for over a minute during our tests.

We note that while all policies are successful in that they can stably walk, adding in dynamic loads significantly widens the sim-to-real gap.
All of the load policies reach significantly lower top speeds on hardware than in simulation. 
While the base walking policy reaches a top speed of \unit[3.0]{m/s} on hardware, undershooting the simulation test by \unit[0.9]{m/s}, the Tray Box policy only reached \unit[1.0]{m/s}, less than half of the simulation test's \unit[2.1]{m/s}.
The Carry Pole policy does even worse, reaching only \unit[0.8]{m/s} compared to \unit[2.5]{m/s} in simulation. 

We see this phenomenon in the general load policies as well. 
Though we see little difference in behavior between the load specific and general policy for the Tray Box load, on the more difficult Carry Pole load, the general load trained from scratch failed where as the bootstrapped one was able to successfully walk. 
This is likely because of all the extra noise that the from scratch policy sees during training. 
The set of training states the from scratch policy experiences is much more varied as it spends the entirety of training seeing states from all loads. 
In contrast, the bootstrapped policy spends over half of its experience only seeing states from the base model, during the training of the initial base policy. 
The extra noise in the training of the from scratch policy likely reduces its ability to transfer, which is fine when the sim-to-real gap is small as for the unloaded and Tray Box case, but when the sim-to-real gap widens like in the Carry Pole load, it fails. 

It is reasonable that adding dynamic loads increases the sim-to-real gap. 
More moving bodies will result in more things to account for and more complex dynamics to model, which all increases the margin of error in our simulation. 
The deficiencies in hardware performance that we experience here indicate that our (and RL's most common) sim-to-real solution of dynamics randomization is not enough for more dynamic tasks like carrying loads. 
This indicates that further research into crossing the sim-to-real gap is necessary.

\section{Conclusion}
In this work we have shown that it is possible to train policies capable of handling dynamic loads while locomoting for human scale bipedal robots.
We find that no additional sensing is required, and that the only modification required from learning normal locomotion policies is to train in the context of the desired loads. 
We also show that it is possible to learn a single general load policy that can handle a number of different dynamic loads while maintaining the walking performance of specialized policies trained on a single specific load. 
In addition, we find that we can bootstrap from a previously trained unloaded walking policy to quickly learn loaded walking policies without loss of performance when compared to policies trained from scratch. 
We demonstrate successful sim-to-real transfer for the load policies, achieving a speed of \unit[0.8]{m/s} while carrying two \unit[5]{kg} swinging weights on each side of the robot.


Our hardware results bring up interesting sim-to-real questions that could be addressed in future work. 
Adding dynamic loads dramatically widens the sim-to-real gap, and it seems like we are getting close to breaking the canonical solution of dynamics randomization. 
It is likely that better strategies are required in order to achieve simulation level of performance on hardware for complex tasks like handling dynamic loads. 
More sophisticated sim-to-real methods like shown in \cite{Mozifian2019} are likely necessary to get full simulation performance on hardware for loaded policies.

\section*{Acknowledgments}
\small{We thank Carl Wilcox for design of hardware for physical load tests, Intel for providing vLab resources and students at the Dynamic Robotics Laboratory for helpful discussions.}

\addtolength{\textheight}{-7cm} 

\def\bibfont{\footnotesize}
\bibliographystyle{IEEEtranN}
\bibliography{loadmass_bib}

\end{document}